\RequirePackage[T1]{fontenc}
\documentclass[a4paper,10pt,conference]{IEEEtran}
\IEEEoverridecommandlockouts
\usepackage{cite}
\usepackage{color}
\usepackage{amsmath,amssymb}
\usepackage{graphicx}
\usepackage[pass]{geometry}
\usepackage{pdflscape}
\usepackage{booktabs,tabularx,array,multirow}
\usepackage{placeins}
\usepackage{balance}
\usepackage[hidelinks]{hyperref}
\graphicspath{{figures/}}
\newcolumntype{L}[1]{>{\raggedright\arraybackslash}p{#1}}
\newcolumntype{Y}{>{\raggedright\arraybackslash}X}
\hypersetup{pdftitle={TeleMultiStream: An Integrated Robot Teleoperation System with Low-Latency Multi-Stream Video Feedback},pdfauthor={Chenchen Lv, Lingchen Hu, Yaosai Zhang, Chunlan Li, Gengkui Wei, Wei Xin, Li Li}}
\title{\LARGE\bfseries TeleMultiStream: An Integrated Robot Teleoperation System with Low-Latency Multi-Stream Video Feedback}
\author{%
\IEEEauthorblockN{Chenchen Lv, Lingchen Hu, Yaosai Zhang, Chunlan Li, Gengkui Wei, Wei Xin, and Li Li\textsuperscript{*}}%
\thanks{First author: Chenchen Lv (\href{mailto:lvbuer@whu.edu.cn}{\texttt{lvbuer@whu.edu.cn}}).}%
\thanks{\textsuperscript{*}Corresponding author: Li Li (\href{mailto:283999281@qq.com}{\texttt{283999281@qq.com}}).}}
\begin{document}
\maketitle
\pagestyle{plain}
\thispagestyle{plain}
\begin{abstract}
Training embodied-intelligence models requires large-scale, high-quality demonstration data. Robot teleoperation provides an important means of collecting these data. Low-latency teleoperation also helps human operators take control promptly and provide corrective demonstrations in DAgger-based interactive learning. This paper presents TeleMultiStream, a robot teleoperation system that combines multi-stream WebRTC video feedback with UDP pose downlink. The system supports heterogeneous video sources and uses buffer reuse, bounded queues, and native XR rendering to reduce processing and waiting. In wired-LAN ZED Mini tests, TeleMultiStream's software and high-performance hardware encoding modes achieved dominant-interval Glass-to-Glass latency medians of 60 and 66~ms, respectively. These medians were 26.8\% and 19.5\% lower than the 82-ms RobotVision-PC reference, respectively. At the lower tested resolutions, the two TeleMultiStream encoding modes provided comparable visual latency. Their respective medians in three-stream Configuration~1 were 72 and 66~ms. In the higher-resolution Configuration~2, high-performance hardware encoding outperformed software encoding in latency and CPU utilization. It reduced mean encoding time from 11.85 to 8.80~ms and the dominant-interval Glass-to-Glass median from 94 to 88~ms. Its CPU utilization was approximately 67\%, compared with 265\% for software encoding. During three-stream reception with high-performance hardware encoding, the mean corrected one-way UDP latencies at 100~Hz were 1.576 and 1.800~ms in Configurations~1 and~2, respectively. Mean estimated packet loss was 0.0033\% in both configurations. These results support responsive teleoperation for robot demonstration collection.
\end{abstract}


\section{Introduction}\label{sec:introduction}

Progress in embodied intelligence increasingly depends on robot models trained with large-scale, high-quality interaction data~\cite{octo,openvla,droid,datascaling,openx}. Teleoperation supplies human demonstrations, whose usefulness depends on interface design and data quality~\cite{rt1,demonstrationmodality,bunnyvisionpro,robomimic,deminf}. DAgger training additionally collects expert corrections at states visited by a learned policy~\cite{dagger}. Human-gated variants let operators take over, demonstrate recovery, and return control to the policy~\cite{hgdagger}. Both data collection and timely intervention require low-latency teleoperation: delayed video obscures the current scene, while delayed commands postpone corrective motion and disturb observation--action alignment~\cite{immersivelatency,onsetlatency,umi}.

Multiple camera views further complicate this requirement. Stereo head cameras provide workspace context and depth cues, while wrist cameras expose contact and local manipulation~\cite{aloha}. Concurrent streams share host computation, network bandwidth, and headset decoding resources. Maintaining responsive feedback therefore requires reducing processing and waiting throughout the video path~\cite{g2gminimization}.

\begin{figure}[!htbp]
  \centering
  \includegraphics[width=\columnwidth]{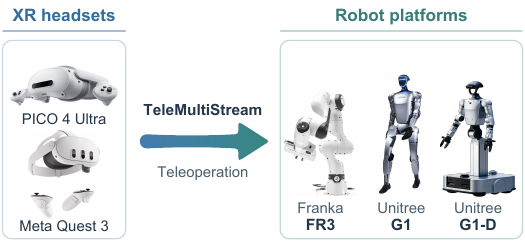}
  \caption{Cross-device teleoperation with TeleMultiStream, connecting XR headsets to robotic arms, humanoids, and mobile manipulators.}
  \label{fig:xr-robot-compatibility}
\end{figure}

\begin{figure}[!htbp]
\centering
\includegraphics[width=0.92\columnwidth,keepaspectratio]{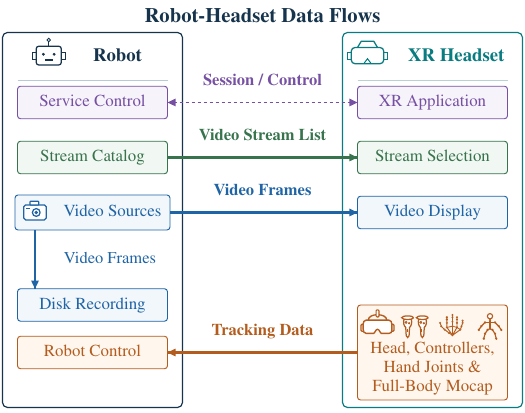}
\caption{Robot--headset data flow: session control, stream discovery, video feedback and recording, and tracking data for robot control.}
\label{fig:system-data-flow}
\end{figure}

To meet these requirements, this paper presents TeleMultiStream, an integrated low-latency robot teleoperation system. Its Streaming Service sends multiple robot-camera views over WebRTC to a native XR receiver, while a UDP downlink carries operator poses to the robot. The system supports teleoperation across heterogeneous XR headsets and robot platforms, as illustrated in Fig.~\ref{fig:xr-robot-compatibility}. Fig.~\ref{fig:system-data-flow} presents the system's video-feedback and pose-downlink data flow.

\begin{samepage}
The main contributions are as follows:\nopagebreak[4]
\begin{enumerate}
\item \textbf{Low-latency multi-stream video feedback.} Source adaptation, buffer reuse, bounded queues, and native XR rendering reduce latency. Common-source dominant-interval Glass-to-Glass medians were 60~ms with software encoding and 66~ms with high-performance hardware encoding, versus 82~ms for RobotVision-PC~\cite{robotvision}.

\item \textbf{Real-time video transport and stage-level latency analysis.} Per-track scheduling over WebRTC supports timely multi-stream transmission~\cite{rfc8835,rfc8834}. Pipeline instrumentation quantifies conversion, encoding, queueing, decoding, and rendering delays, enabling detailed analysis of where latency accumulates.

\item \textbf{Flexible video sources and encoding backends.} Common interfaces combine RealSense, ZED-M, and raw or compressed ROS image inputs, including compatible encoded passthrough. Cross-platform software encoding and platform-specific hardware backends adapt the system to different camera layouts, host resources, and deployment platforms.
\end{enumerate}
\end{samepage}

The evaluation uses PICO 4 Ultra and PICO Swan headsets. It measures Glass-to-Glass video latency, local processing times, CPU utilization, and the GPU/SoC-rail power change during video transmission. It also assesses corrected one-way pose latency and estimated packet loss during concurrent video reception. By integrating responsive pose delivery with low-latency multi-stream visual feedback, TeleMultiStream provides a reusable foundation for robot demonstration collection and the development of embodied-intelligence applications.

Section~\ref{sec:background} distinguishes the relevant formats, protocols, and reference implementations. Section~\ref{sec:architecture} describes the video sender and receiver, timing instrumentation, and UDP pose downlink. Section~\ref{sec:experiments} reports video comparisons and concurrent pose-downlink measurements. Section~\ref{sec:conclusion} summarizes the system's capabilities and measured performance.

\section{Background and Related Work}\label{sec:background}

\subsection{Related Teleoperation Systems}

XRoboToolkit~\cite{xrobotoolkit} provides an OpenXR-based framework
integrating tracking, stereoscopic feedback, and robot control, with
adoption in NVIDIA's GR00T whole-body teleoperation workflow and TWIST2
demonstrating its value as reusable infrastructure for robot data
collection~\cite{nvidia_gr00t_teleop,twist2}.
Open-TeleVision~\cite{television} combines real-time stereoscopic video
with head-controlled active vision through a web-based interface,
providing an open visual-feedback implementation that has supported
downstream development such as Unitree's XR teleoperation
stack~\cite{unitree_xr_teleoperate}.

However, their reported video-feedback evaluations primarily concern
a single stereo source: XRoboToolkit reports aggregate Glass-to-Glass
latency, while Open-TeleVision emphasizes active vision and
imitation-learning performance. These evaluations do not provide a
detailed conversion-to-rendering latency breakdown for concurrent
head- and wrist-camera streaming, leaving the effects of shared
computation, network bandwidth, and headset decoding capacity
insufficiently characterized. TeleMultiStream addresses this gap
through stage-level latency measurement and coordinated sender--receiver
pipeline optimization, targeting low-latency multi-camera feedback
within an integrated XR teleoperation system.

\subsection{Video Coding and Transport}

TeleMultiStream uses H.264 for compatibility with Jetson encoding and PICO hardware decoding, and WebRTC for multi-track RTP delivery~\cite{rfc6184,rfc8835}. WebRTC negotiates tracks through Session Description Protocol (SDP) and establishes connectivity through Interactive Connectivity Establishment (ICE). Low latency still depends on conversion, encoder buffering, reception queues, decoding, and rendering~\cite{gcc}.

Software encoding uses CPU-based libx264 with configurable group-of-pictures (GOP) structure and buffering. Multiple high-resolution streams increase CPU demand. Jetson hardware encoding uses GStreamer: VIC converts images to NV12 in NVIDIA multimedia memory (NVMM), and NVENC performs H.264 compression. Conversion, memory import, and queueing remain part of the hardware path, so engine frequencies and complete-path latency must be considered alongside encoder execution time.

\subsection{Reference Sender Implementations}

\begin{figure}[!htbp]
\centering
\resizebox{0.85\columnwidth}{!}{\input{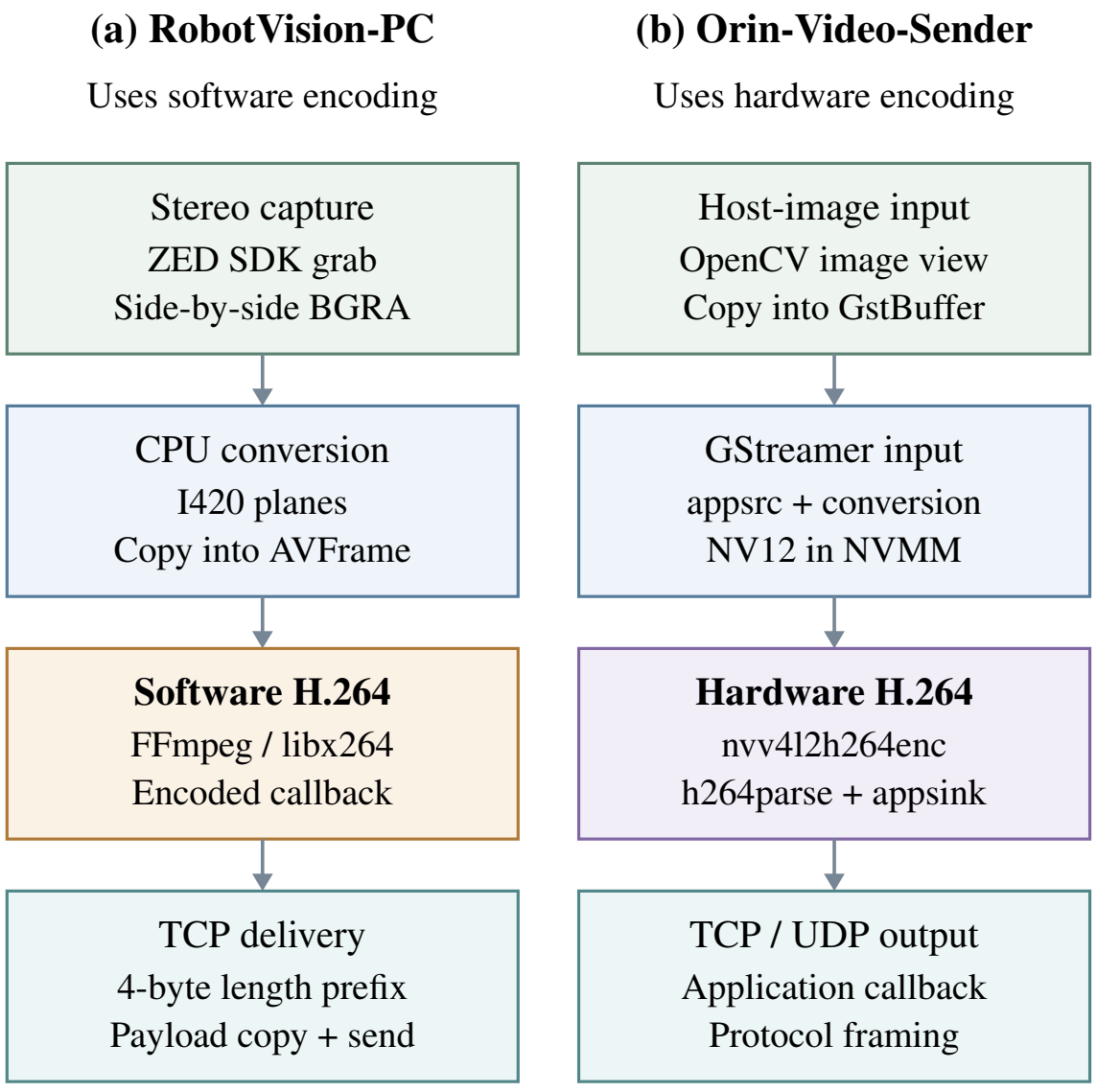}}
\caption{Sender pipelines implemented in XRoboToolkit. Panel (a) shows software H.264 encoding with FFmpeg/libx264 in RobotVision-PC. Panel (b) shows hardware H.264 encoding with \texttt{nvv4l2h264enc} in Orin-Video-Sender. Arrows show the order of conversion, copying, and delivery.}
\label{fig:baselines}
\end{figure}

Fig.~\ref{fig:baselines} contrasts the software and hardware paths of two XRoboToolkit senders~\cite{robotvision,orinsender}.

RobotVision-PC~\cite{robotvision} converts ZED-M side-by-side images to I420, copies them into FFmpeg frames, and uses libx264 with \texttt{ultrafast}, \texttt{zerolatency}, and no B frames. Encoded output is copied into a length-prefixed buffer for synchronous TCP delivery. These boundaries motivate TeleMultiStream's direct conversion, shared-buffer ownership, and independent output queues.

Orin-Video-Sender copies ZED-M BGRA images into GStreamer buffers, converts them to NV12 in NVMM, and encodes them with \texttt{nvv4l2h264enc}. The default output uses TCP; a separate implementation supports UDP. This path illustrates the conversion, copying, and queueing costs that remain with hardware encoding.

RobotVision-PC is the quantitative reference for the complete video path; Orin-Video-Sender is an architectural reference for Jetson hardware encoding.

\section{System Architecture}\label{sec:architecture}

\subsection{System Overview}

\begin{figure*}[!t]
\centering
\includegraphics[width=0.90\textwidth,keepaspectratio]{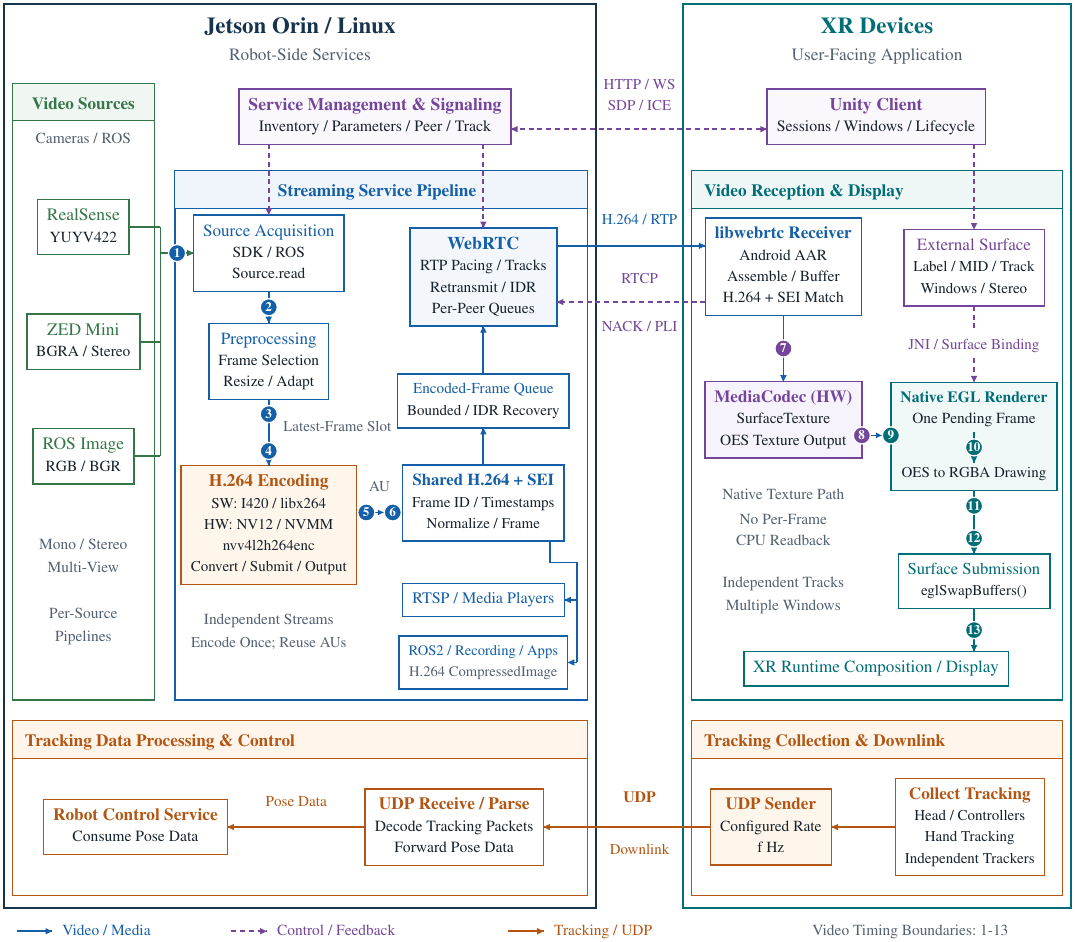}
\caption{TeleMultiStream architecture, using Jetson Orin running Linux as an example robot-side host. Blue arrows denote multi-stream video feedback to the XR operator interface, purple dashed arrows video-session control and media feedback, and orange arrows UDP pose downlink to the robot. Pose samples are parsed on the robot-side host and forwarded to the robot control service. Numbers 1--13 identify video-processing boundaries.}
\label{fig:architecture}
\end{figure*}

TeleMultiStream connects the robot and the XR operator interface through low-latency multi-stream video feedback, complemented by UDP pose downlink (Fig.~\ref{fig:architecture}). The Streaming Service sender supports different host architectures. Fig.~\ref{fig:architecture} illustrates a Jetson Orin host running Linux. Video sources feed a common pipeline that converts camera inputs into encoded media, negotiates stream parameters, schedules transmission, and records processing stages. The native PICO receiver performs media reception, hardware decoding, and low-latency Surface rendering before XR composition. The sender limits image processing and network backlog. The receiver limits copying and waiting during display handoff.

Video-session control and video media are logically separate. HTTP exposes the video inventory, stream state, and start/stop operations. WebSocket carries WebRTC negotiation. After negotiation, H.264 is transmitted continuously over RTP, while RTCP communicates retransmission and refresh requests. The same encoded output can also be published to ROS2 for recording or downstream processing. TeleMultiStream also supports optional Real-Time Streaming Protocol (RTSP) output for media players~\cite{rfc7826}. Section~\ref{sec:experiments} evaluates the WebRTC video path of TeleMultiStream.

Along the video-feedback path, each processing boundary has explicit input, output, and ownership rules. Pixel buffers remain referenced by their consumers, waiting queues are bounded, and frame metadata is attached at a common encoded-output boundary before fan-out. The native PICO receiver uses a camera label to select the correct track and display window, and generation identifiers to distinguish replaced connections or Surfaces. These mechanisms support independent multi-window display and stage-level diagnosis without requiring Unity's main thread to handle every frame.

The pose downlink carries head, controller, hand, full-body, and independent-tracker data from the XR operator interface to the robot-side host. A separate receiver parses these samples and forwards pose data to the robot control service. Section~\ref{sec:pose-downlink} describes the pose format and its use over local UDP and public-network control channels.

\subsection{Multi-Stream Video Sender}\label{sec:video-sender}

Each video source runs in its own processing pipeline, producing an encoded stream through an encoder or compatible encoded passthrough. The pipeline shares that output between its network and ROS2 branches. Each subscribed stream maps to a negotiated video track. Tracks for one endpoint share a connection, and camera labels bind them to display windows. Raw-frame slots and output queues belong to individual source pipelines, while transmission queues belong to individual tracks within each connection.

\subsubsection{Source Adaptation and Buffer Management}

Streaming Service supports native RealSense and ZED-M cameras and ROS raw or compressed image topics. Its input interfaces accept RGB/BGR, BGRA, YUYV, and I420 raw images, or JPEG and compatible H.264 data. Table~\ref{tab:pixel-formats} summarizes the sources, formats, and adaptation paths. Separate raw and encoded interfaces allow compatible compressed input to bypass decoding and re-encoding.

\begin{table}[!t]
\centering
\caption{Video sources and adaptation paths}
\label{tab:pixel-formats}
\small
\setlength{\tabcolsep}{3pt}
\renewcommand{\arraystretch}{1.08}
\begin{tabularx}{\columnwidth}{@{}L{.25\columnwidth}Y@{}}
\toprule
\textbf{Source} & \textbf{Format / handling} \\
\midrule
RealSense & Reuses YUYV422 SDK buffers. Adapts to I420 (SW) or NV12 (HW). \\
ZED-M & Adapts BGRA32 left, right, or stereo views using a CPU buffer pool. \\
ROS raw image & Reuses compatible \texttt{rgb8}/\texttt{bgr8} buffers. Converts or resizes as needed. \\
ROS compressed & Forwards or transcodes JPEG and passes through compatible H.264. \\
\bottomrule
\end{tabularx}
\par\vspace{3pt}
\begin{minipage}{\columnwidth}
\footnotesize Raw inputs include RGB24/BGR24, BGRA32, YUYV422, and I420/YUV420p. Compatible I420 buffers can be reused. JPEG may be forwarded as MJPEG or transcoded to H.264. Native modes depend on the device/SDK. NV12/NVMM is internal to HW encoding.
\end{minipage}
\end{table}

Adapters preserve buffer ownership and stride, reuse compatible storage, and convert only when required. Within each source pipeline, acquisition and encoding share one waiting raw-frame slot: a newer frame replaces stale pending input while the frame being processed remains valid. H.264 passthrough additionally preserves prediction dependencies and decoder state.

\subsubsection{Software and Hardware Encoding}

\begin{figure}[!htbp]
\centering
\resizebox{0.85\columnwidth}{!}{\input{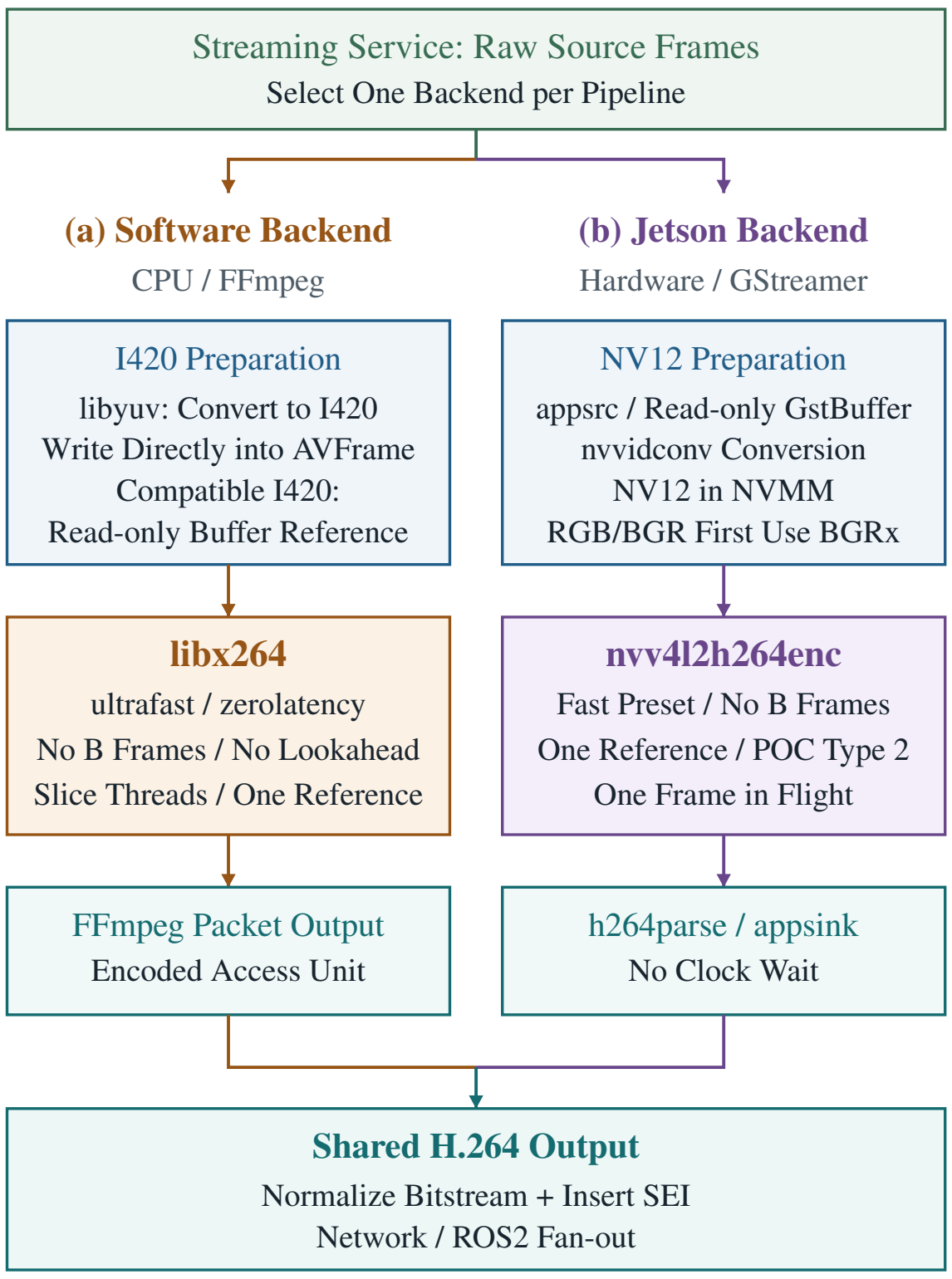}}
\caption{Streaming Service encoding backends on Jetson Orin. Panel (a) shows I420 input to the libx264 software encoder. Panel (b) shows NV12 input in NVMM for \texttt{nvv4l2h264enc}. Both backends share bitstream normalization and supplemental enhancement information (SEI) insertion. Baseline and high-performance hardware encoding use the same panel (b) pipeline. Buffer reuse limits host copies, although conversion and hardware-memory import still incur work.}
\label{fig:encoders}
\end{figure}

Streaming Service selects an encoding backend according to the capabilities of the robot-side host. The Jetson Orin example uses libx264 software encoding or Jetson hardware encoding (Fig.~\ref{fig:encoders}). Both accept adapted source frames and produce encoded access units for a shared output stage. Their implementation is evaluated at the boundaries where data enters the encoder, waits internally, and becomes available to the application.

For software encoding, libyuv converts non-I420 input directly into an FFmpeg \texttt{AVFrame}. Compatible I420 buffers use read-only references. Conversion, codec submission, and packet retrieval are timed separately. libx264 uses \texttt{ultrafast} and \texttt{zerolatency}, disables lookahead and B frames, and combines slice threads with one reference picture to limit buffering.

Disabling scene-cut keyframes makes refresh scheduling predictable. A fixed GOP or an external schedule provides periodic instantaneous decoding refresh (IDR) pictures. New connections and decoder recovery can also trigger an IDR request. Periodic IDRs can be staggered across streams to reduce simultaneous bursts. Shorter GOPs ease joining and recovery but increase intra-frame traffic. Optional Video Buffering Verifier (VBV) constraints limit coded-rate variation, but packet pacing and network adaptation remain separate. These settings favor responsiveness. Comparable image quality can therefore require more bandwidth.

For Jetson hardware encoding, compatible YUYV/BGRA images enter \texttt{appsrc} through read-only \texttt{GstBuffer} references that retain source ownership. The \texttt{nvvidconv} element converts these images to NV12 in NVMM. RGB24/BGR24 input first passes through \texttt{videoconvert} for conversion to BGRx. Reference reuse avoids selected host copies but does not remove conversion or hardware-memory import.

Jetson encoding uses \texttt{nvv4l2h264enc} with \texttt{preset-level=1}, no B frames, one reference picture, \texttt{poc-type=2}, VUI, and SPS/PPS insertion. The fast preset targets shorter processing, \texttt{poc-type=2} reduces reordering, and parameter sets support decoder startup. The existing pipeline also sets \texttt{maxperf-enable=1}, but this flag is not used as evidence of maximum engine frequency on JetPack~6. Frequency control is described below. \texttt{h264parse} produces access units, while \texttt{appsink} disables clock waiting and limits buffering. One frame remains in flight, with no extra queue after \texttt{appsrc}.

Both backends share bitstream normalization and SEI insertion before network and ROS2 fan-out, preserving common frame identity and measurement boundaries. Their complete paths determine the latency comparison.

\subsubsection{Session Management and Transmission Scheduling}

Runtime settings control resolution, frame rate, bitrate, encoder, GOP, and send budget. The receiver discovers sources over HTTP, negotiates H.264 tracks using SDP over WebSocket, and establishes connectivity through ICE. Tracks share one connection per endpoint; camera labels and negotiated media identifiers (MIDs) associate them with display windows. Multi-track transport does not synchronize camera exposure, and display layout does not change encoded frame geometry.

When negotiated, the RTP playout-delay extension uses 10-ms units. Zero minimum and maximum request prompt presentation without guaranteeing zero latency. The receiver retains libwebrtc buffering, while the Surface renderer avoids an extra fixed playback delay.

For each source pipeline, the network and ROS2 output branches share encoded access units and retain one and eight waiting frames, respectively, excluding the frame in transmission. Independent bounded queues limit backlog without encoding again for each consumer.

After a predictive H.264 frame is discarded, the affected path skips dependent pictures until an IDR and requests encoder recovery. A negative acknowledgement (NACK) requests retransmission of a missing RTP packet. A picture loss indication (PLI) requests a prediction refresh. The sender coalesces and rate-limits refresh requests, retains pending requests, and can prioritize an initial IDR for a new receiver.

Downstream of the per-pipeline output queues, each track in each remote connection has an independent worker, transmission queue, and retransmission cache. Original and retransmitted packets share a pacing budget. The sender bounds both queue length and packet waiting time. Compatible frame rates and GOPs permit staggered periodic IDRs without delaying startup or recovery. The budget includes estimated protocol overhead, and phase staggering does not hold frames while waiting for another camera. Upstream encoded passthrough cannot be scheduled this way.

The ROS2 branch publishes shared H.264 access units with frame identity and timestamps, avoiding re-encoding but retaining payload copying and DDS serialization. Best-effort and reliable QoS are supported; reliability can add waiting or backpressure. Subscribers must handle parameter sets and IDR recovery.

\subsubsection{Build and Hardware Frequency Conditions}

Build optimization and device operating state can affect measured timing. Software conversion and encoding use CPU resources. By contrast, the hardware path uses VIC for conversion and NVENC for compression. These conditions should therefore be distinguished from the encoding backend.

High-performance hardware encoding (HW-HP) preserves the sender/receiver code and GStreamer pipeline while setting VIC and NVENC device-frequency nodes to supported maximum frequencies; baseline hardware encoding (HW) uses automatic frequency management. NVIDIA recommends device-node configuration because the encoder maximum-performance mode is deprecated in JetPack~6~\cite{jetpack6frequency}. The measured frequency changes, rather than \texttt{maxperf-enable=1} alone, establish the HW-HP operating state.

VIC conversion and NVENC encoding dominate the HW-HP encoding time for the tested ZED-M streams. Together, they account for approximately 97--98\% of the measured total. The comparison therefore focuses on these two stages and the directly measured total. Minor preparation and submission overheads are therefore omitted. In addition, encoder output wait overlaps asynchronous conversion and encoding. These overlapping intervals must not be added.

\subsection{Native PICO Video Receiver}\label{sec:video-receiver}

\subsubsection{Native Reception and Hardware Decoding}

The custom Android AAR, built on libwebrtc M114, keeps reception, decoding, and drawing in the native path. Unity C\# uses the Java Native Interface (JNI) to control source selection, sessions, and windows. Only commands and diagnostic snapshots cross this boundary. ICE, DTLS, RTP reception, and hardware-decoder integration remain libwebrtc functions. The AAR extends decoder interfaces, track management, Surface rendering, and frame association, keeping per-frame pixel arrays out of managed memory.

The H.264 decoder factory selects hardware-accelerated Android decoders and reports unsupported configurations instead of silently falling back to CPU decoding. A wrapper extracts SEI, associates frames, records decoder events, and verifies GPU \texttt{TextureBuffer} output without automatic \texttt{toI420} conversion. Unexpected decoder output is recorded explicitly, while the underlying decoder remains responsible for reconstructing coded pictures.

\subsubsection{Surface Rendering and Frame Scheduling}

Compatible EGL contexts and external OES textures let \texttt{VideoFrameDrawer} sample decoded images directly into the PICO Android Surface. \texttt{eglSwapBuffers} submits the result for XR composition, removing CPU readback, JNI pixel transfer, and Unity texture re-upload. After submission, GPU work, composition, and scanout can continue before the frame appears on the display.

Each track has its own render thread and one pending frame. New arrivals replace stale waiting references, while frames being drawn remain valid. The renderer draws frames as they become available. Its optional rate and age limits operate independently of the sender and the RTP jitter buffer.

After PICO creates an External Surface, JNI binds it to the labeled track. First-frame readiness is reported only after successful native submission. Viewports use the actual Surface dimensions, and window aspect ratios follow the decoded geometry. The layout can change within an existing connection.

Labels, MIDs, tracks, and decoders remain associated. Connection and Surface generations reject stale callbacks, and native rendering is unbound before Unity releases its target. Multiple tracks decode and display independently, subject to hardware decoder capacity, full-frame resolution, frame rate, and bandwidth.

\subsubsection{Stereo Views and Window Lifetime}

Side-by-side stereo places both camera views in one encoded frame, sharing an access unit and media time and requiring only one decoding operation. The XR compositor samples the corresponding half of the decoded Surface for each eye, preserving binocular disparity without per-frame managed cropping or duplicate splitting. Sampling order must match the source; stereo accuracy also depends on camera acquisition and calibration.

Windows share a session but retain independent Surfaces and layout. Hiding or pausing can preserve reception and decoding for fast resume while transmission continues. Native unbinding precedes target release, and generation checks reject obsolete events. A stalled track is diagnosed locally before restarting the whole connection.

\subsection{Video Timing and Instrumentation}\label{sec:video-timing}

\subsubsection{Frame Identity and Clock Domains}

After \texttt{Source.read} returns a valid frame, the pipeline assigns \texttt{frameId} and samples \texttt{captureSystemUs}. Frame identifiers restart at one for each pipeline. Table~\ref{tab:sei-fields} maps the source-output, encoded-output, and pre-SEI timestamps to boundaries 2, 5, and 6 in Fig.~\ref{fig:architecture}, respectively. The sender records the submission timestamp before SEI rewriting, output queueing, and network transmission.

Local durations use monotonic time. Cross-device metadata uses Unix microseconds. Media PTS and DTS use a separate time base. A cross-device comparison uses timestamps in a common, aligned clock domain.

\begin{table}[!t]
\centering
\caption{Frame identity and SEI timestamps}
\label{tab:sei-fields}
\small
\setlength{\tabcolsep}{3pt}
\renewcommand{\arraystretch}{1.08}
\newcommand{\stagepoint}[1]{\begingroup\fontencoding{OMS}\fontfamily{cmsy}\fontsize{11}{11}\selectfont\textcircled{\normalfont\fontsize{6}{6}\selectfont #1}\endgroup}
\newcommand{\stagepair}[2]{\mbox{\stagepoint{#1}\,{\fontsize{9}{9}\selectfont$\rightarrow$}\,\stagepoint{#2}}}
\begin{tabularx}{\columnwidth}{@{}L{.51\columnwidth}Y@{}}
\toprule
\textbf{Field} & \textbf{Meaning / event} \\
\midrule
\texttt{frame\_id} & Same-frame identity. \\
\texttt{capture\_system\_us} & Valid-frame return \stagepoint{2}. \\
\texttt{encode\_done\_system\_us} & Encoded access unit ready \stagepoint{5}. \\
\texttt{sender\_submit\_system\_us} & Before SEI rewrite and output queueing \stagepoint{6}. \\
UUID / version / flags & FTM1 record validation. \\
\bottomrule
\end{tabularx}
\par\vspace{3pt}
\begin{minipage}{\columnwidth}
\footnotesize Timestamps use Unix microseconds. Boundary \stagepoint{2} marks valid-frame return from the SDK or source interface. Boundary \stagepoint{6} precedes RTP transmission. Numbers refer to Fig.~\ref{fig:architecture}.
\end{minipage}
\end{table}

An H.264 \texttt{user\_data\_unregistered} SEI record uses a dedicated UUID and FTM1 header for version, validity, length, and frame data. The common SEI stage precedes fan-out, so the independent network and ROS2 branches carry the same frame identity and timestamps while retaining separate output queues and delivery schedules. The record does not change pixels. The custom receiver parses SEI before decoding. Other decoders can ignore this metadata.

\subsubsection{Local Stage Timing and Diagnostics}

\begin{table}[!t]
\centering
\caption{Timing boundaries along the video path}
\label{tab:sender-intervals}
\small
\setlength{\tabcolsep}{3pt}
\renewcommand{\arraystretch}{1.08}
\newcommand{\stagepoint}[1]{\begingroup\fontencoding{OMS}\fontfamily{cmsy}\fontsize{11}{11}\selectfont\textcircled{\normalfont\fontsize{6}{6}\selectfont #1}\endgroup}
\newcommand{\stagepair}[2]{\mbox{\stagepoint{#1}\,{\fontsize{9}{9}\selectfont$\rightarrow$}\,\stagepoint{#2}}}
\begin{tabularx}{\columnwidth}{@{}Yr@{}}
\toprule
\textbf{Stage} & \textbf{Boundary} \\
\midrule
\multicolumn{2}{@{}l}{\textit{Sender}} \\
Source read & \stagepair{1}{2} \\
Preprocessing & \stagepair{2}{3} \\
Raw-frame queueing & \stagepair{3}{4} \\
Encoding time & \stagepair{4}{5} \\
\quad Encoder output wait & Within \stagepair{4}{5} \\
Sender processing time & Outer send-call return\textsuperscript{a} \\
\midrule
\multicolumn{2}{@{}l}{\textit{Output / reception}} \\
Encoded access unit to SEI entry & \stagepair{5}{6} \\
Pre-SEI timestamp to decoder input & \stagepair{6}{7} \\
\midrule
\multicolumn{2}{@{}l}{\textit{Receiver}} \\
Decoding time & \stagepair{7}{8} \\
Frame handoff & \stagepair{8}{9} \\
Render queueing & \stagepair{9}{10} \\
CPU drawing & \stagepair{10}{11} \\
Surface swap & \stagepair{12}{13} \\
\bottomrule
\end{tabularx}
\par\vspace{3pt}
\begin{minipage}{\columnwidth}
\footnotesize Numbers refer to Fig.~\ref{fig:architecture}. For a matched frame on aligned clocks, instrumented pipeline latency spans valid-frame return to Surface-swap return (\stagepair{2}{13}). It includes queues, transport, and receiver buffering in \stagepair{6}{7}. Encoder output wait is part of encoding time. \textsuperscript{a}Sender processing time is a composite local interval with an unnumbered endpoint. XR composition and scanout follow Surface submission.
\end{minipage}
\end{table}

Table~\ref{tab:sender-intervals} identifies the timestamp boundaries used for local-stage timings. Encoding time includes conversion and encoder output wait. Sender processing time is a composite interval that ends when the outer send call returns, which can precede completion of RTP submission. Adding a composite interval to its components would count some work twice.

The decoder wrapper associates decoder instance, media-time markers, texture identity, track, and session to match input with output. Latency estimates are marked as unavailable when SEI metadata are missing or frame association fails. They are not set to zero.

The receiver records decoding time (7--8), frame handoff (8--9), render queueing (9--10), CPU drawing (10--11), and Surface swap (12--13). CPU drawing ends at draw-call return. Surface-swap time is measured from the invocation of \texttt{eglSwapBuffers} until its return.

Receiver checks cover SEI validity, decoder state, and continuity across source labels, connection/pipeline generations, and frame IDs. Sender counters, RTP statistics, and receiver events help distinguish network loss from frame filtering, replacement, or decoder recovery. SEI associates frames but does not verify pixel integrity.

RTCStats and native renderer events distinguish reception from post-decode waiting, although some internal queues remain unobservable. Unity retrieves native events in batches for each connection. Where possible, JNI and parsing run off the Unity main thread. The user interface updates at most once per second. Disabling diagnostics bypasses wrapping, SEI scanning, and association to assess instrumentation overhead.

\subsubsection{Cross-Device Alignment and Latency Boundaries}

Receiver-local durations use a monotonic clock. Cross-device estimates require matched frame identity and aligned system clocks~\cite{communicationlatency}. Let $t_0$ and $t_1$ denote PICO system-clock timestamps at request transmission and reply reception, respectively, and let $T_{\mathrm{robot}}$ denote the robot system-clock timestamp carried in that reply. All three values are expressed in microseconds since the Unix epoch, each in its device's clock domain. With no clock step during the exchange, an approximately symmetric round trip with small processing delay yields
\begin{equation}
\widehat{\Delta}=T_{\mathrm{robot}}-\frac{t_0+t_1}{2}.
\label{eq:clock-offset}
\end{equation}
Here, $\widehat{\Delta}$ estimates robot time minus PICO time. Adding it to a PICO system-clock timestamp expresses that timestamp in the robot clock domain under the assumptions above.

Glass-to-Glass latency measures the time from a change in the physical scene to its appearance on the headset display. Within this path, instrumented pipeline latency (\texttt{captureToSurface}) starts when the camera SDK or source interface returns a valid frame and ends when \texttt{eglSwapBuffers} returns on PICO (boundaries 2--13 in Fig.~\ref{fig:architecture}). For a matched frame, the difference between these timestamps on aligned clocks covers conversion, encoding, transmission, decoding, rendering, and all intervening waits. It can also be partitioned using adjacent timestamps for that frame. For example, boundaries 6--7 include SEI preparation, sender queues, packetization, transport, and receiver buffering.

Two intervals lie outside the instrumented pipeline. The first runs from the scene change through camera acquisition to valid-frame return. The second runs from Surface-swap return through XR composition and display scanout until the frame becomes visible. External optical measurements capture the complete path and are subject to sampling and timing resolution.

\subsection{UDP Pose Downlink}\label{sec:pose-downlink}

The pose downlink carries operator motion to the robot control service. Its dedicated sending thread has a configurable frequency. The recommended range is 50--500~Hz, depending on the application and transport capacity. Missed sampling periods do not trigger bursts of stale samples. The same pose representation supports local UDP delivery and public-network real-time communication (RTC) control messaging. A separate functional test confirmed pose delivery through Tencent Cloud TRRO.

The bitmask-based compact binary protocol uses a 5-byte header containing a one-byte \texttt{sourceMask} and a signed 32-bit millisecond timestamp. The head block comes first, followed in order by controller poses and inputs, hand joints, full-body joints, and independent trackers. Existing coordinate, pose-component, and joint-order conventions are preserved. Each pose contains three position components and a quaternion, for a total of seven 32-bit floats (28~B). Multibyte values use big-endian order without padding. Unselected blocks consume no space.

Within the currently supported source combinations, a complete sample remains below 1~kB when all available data except dual-hand joints are selected. This fits the 1~kB per-message limit of RTC control channels such as Tencent Cloud TRRO's high-frequency non-reliable mode~\cite{trrocontrol}. Even without any other pose data, the $2\times26$ hand joints alone require 1456~B (52 poses at 28~B each), excluding the 5-byte header. Such samples therefore require application-level splitting and receiver reassembly when carried over a 1~kB control channel.

\FloatBarrier
\section{Experiments and Results}\label{sec:experiments}

The evaluation compares TeleMultiStream with RobotVision-PC~\cite{robotvision} using Glass-to-Glass latency. It also compares software encoding (SW), baseline hardware encoding (HW), and high-performance hardware encoding (HW-HP). Local-stage timings, CPU utilization, rail power, and estimated module power characterize engine-frequency and stream-load effects. Separate UDP trials measure corrected one-way pose latency and estimate packet loss during concurrent video reception.

\subsection{Experimental Conditions}

\begin{figure}[!htbp]
\centering
\begin{minipage}[t]{0.48\columnwidth}
\centering
\includegraphics[width=\linewidth,height=0.155\textheight,keepaspectratio]{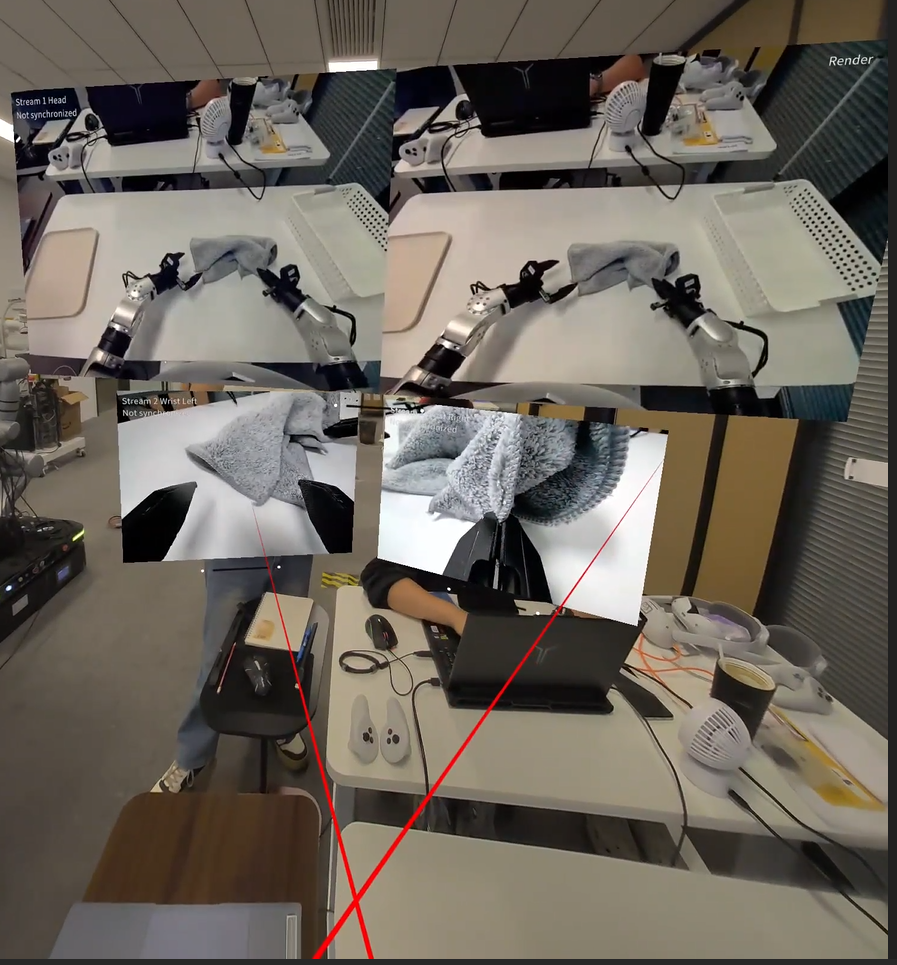}
\par\vspace{2pt}
{\footnotesize (a) First-person view}
\end{minipage}\hfill
\begin{minipage}[t]{0.48\columnwidth}
\centering
\includegraphics[width=\linewidth,height=0.155\textheight,keepaspectratio]{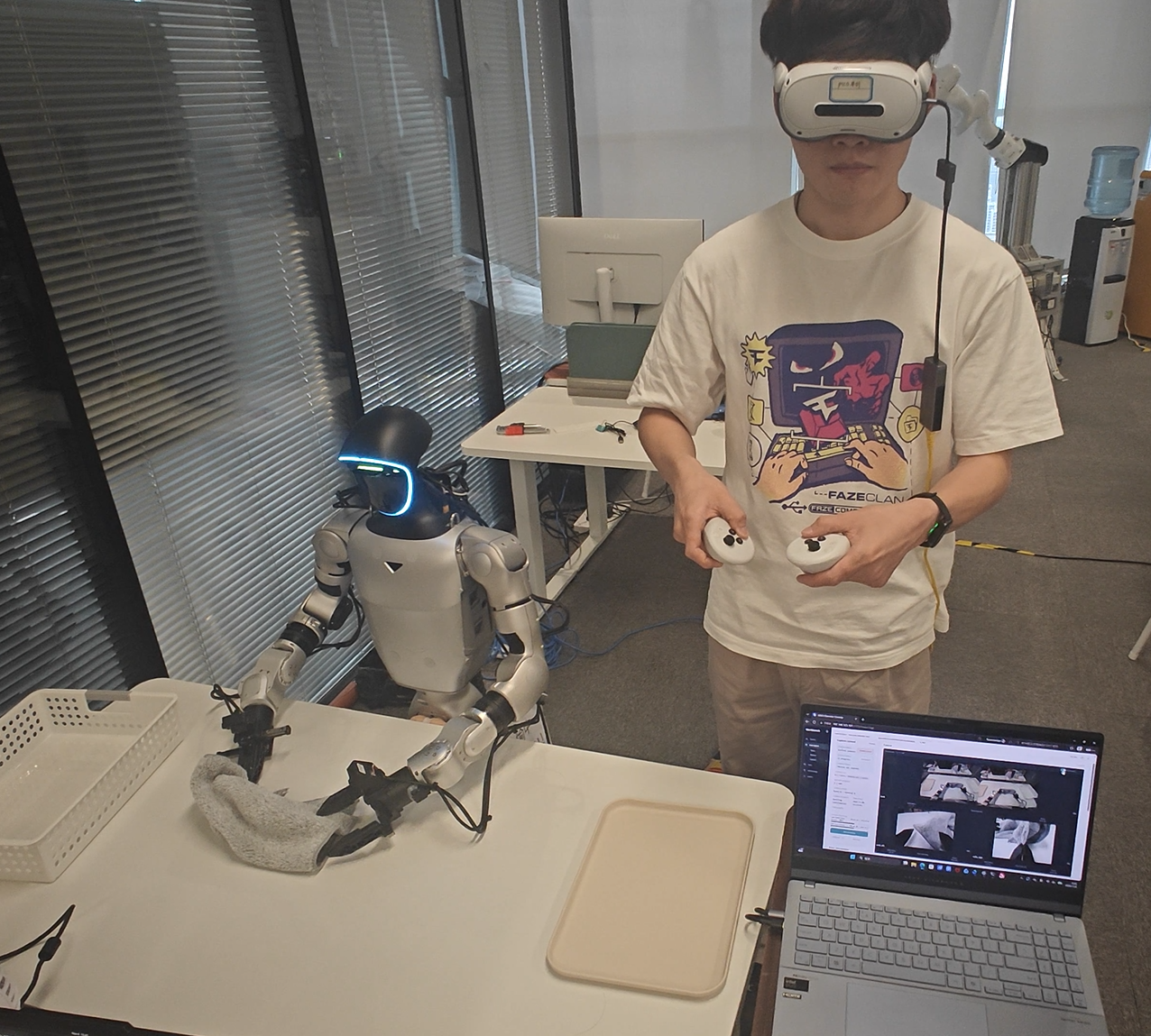}
\par\vspace{2pt}
{\footnotesize (b) Third-person view}
\end{minipage}
\caption{Teleoperation during demonstration collection. (a) XR display of head- and wrist-camera streams. (b) Operator using the headset and controllers.}
\label{fig:teleop-use}
\end{figure}

The default platform used an NVIDIA Jetson AGX Orin 64GB host on a wired LAN. The default receiver was PICO 4 Ultra (Fig.~\ref{fig:teleop-use}). RobotVision-PC and TeleMultiStream used the same host and network. The Configuration~2 software and Hikvision tests also used PICO Swan. Physical connection constraints limited the Hikvision camera to an x86 desktop. Its tests therefore used software encoding.

The video evaluation comprises 35 tests. These include 10 common-source tests, 9 Configuration~1 tests, 10 Configuration~2 tests, and 6 Hikvision tests. Table~\ref{tab:exp-config} gives the source settings and repeats per condition. The pose-downlink evaluation comprises 30 trials. It uses five trials per configuration/encoder combination, covering Configurations~1 and~2 with SW, HW, or HW-HP.

\begin{table}[!t]
\centering
\caption{Sources and conditions of the experiments}
\label{tab:exp-config}
\small
\setlength{\tabcolsep}{3pt}
\renewcommand{\arraystretch}{1.08}
\begin{tabularx}{\columnwidth}{@{}L{.28\columnwidth}Y@{}}
\toprule
\textbf{Setting} & \textbf{Video / comparison} \\
\midrule
Common ZED-M\newline 60 FPS, 1 stream & ZED-M: $2560\times720$ / 1 Mbps.\newline RobotVision-PC: 3 runs.\newline SW/HW: 2 runs each. HW-HP: 3 runs. \\
Config. 1\newline 30 FPS, 3 streams & ZED-M: $2560\times720$ / 8 Mbps.\newline 2 RS: $640\times480$ / 4 Mbps each.\newline 4 Ultra. SW/HW/HW-HP: 3 runs each. \\
Config. 2\newline 30 FPS, 3 streams & ZED-M: $3840\times1080$ / 30 Mbps.\newline 2 RS: $1280\times720$ / 8 Mbps each.\newline SW: 2 runs per headset. HW/HW-HP: 3 runs each on 4 Ultra. \\
Hikvision\newline 30 FPS, 1 stream & $8192\times3072$ / 80 Mbps.\newline x86 host, SW only: 3 runs per headset. \\
\bottomrule
\end{tabularx}
\par\vspace{3pt}
\begin{minipage}{\columnwidth}
\footnotesize RS denotes RealSense. Bitrates are targets. ZED-M and Hikvision resolutions refer to full side-by-side stereo frames.
\end{minipage}
\end{table}

An external synchronized ZED-M stereo camera simultaneously recorded the physical LED panel and its headset image. The recordings yielded 802 valid Glass-to-Glass observations. LED increments were 2~ms, or 5~ms for Hikvision. Multi-stream measurements used the ZED-M view while two RealSense streams provided concurrent load.

Approximate sender CPU utilization was read from Linux \texttt{top}. Power was measured on \texttt{VDD\_GPU\_SOC}, the GPU/SoC rail supplying NVENC and VIC~\cite{orinpower}.

\subsection{Measurement Boundaries and Interval Selection}

An external camera measures Glass-to-Glass latency from the physical scene to the headset display~\cite{g2gmeasurement}. Section~\ref{sec:video-timing} defines the boundaries of instrumented pipeline latency. The local-stage tables use separate sender and receiver windows to compare where processing time was spent.

For each combination of source, configuration, encoder, and headset, the $N$ valid Glass-to-Glass observations were sorted. The dominant interval was the narrowest contiguous interval containing at least $k$ observations, where
\begin{equation}
k=\lceil0.80N\rceil.
\label{eq:minimum-coverage}
\end{equation}
For equal-width intervals, selection favored the midpoint closest to the full-sample median and then the smaller lower bound. All observations within the selected bounds were included, including boundary ties. The resulting $M$ observations determine the dominant-interval median. Actual interval coverage is
\begin{equation}
p_{\mathrm{actual}}=\frac{M}{N}\times100\%.
\label{eq:actual-coverage}
\end{equation}
Full-sample statistics retain all $N$ observations, including those outside the dominant interval.

Local-stage comparisons give each test mean equal weight. Differences were calculated before rounding. The sender's 30-frame windows and the receiver's 1024-frame windows represent different populations. Independent repetition is counted by test rather than by frame or overlapping window. HW-HP stage means use the same tests as the corresponding comparison groups. Cross-device log estimates remain diagnostic because residual clock-offset errors can affect their absolute values. External Glass-to-Glass measurements provide the visual-latency results.

\subsection{Glass-to-Glass Latency Results}

Table~\ref{tab:e2e-summary} summarizes Glass-to-Glass latency across video conditions. The common ZED-M rows compare complete implementations on the same host and network. Panel (a) reports the dominant interval, coverage, and median; panel (b) gives the full-sample range and mean in the same row order.

The relatively broad observed ranges partly reflect the finite frame rates in the current setup. A scene change can occur between camera frames, and a decoded frame can arrive between display refreshes. The waiting time depends on the relative phase of these cycles~\cite{g2gmeasurement}. The synchronized ZED-M measurement camera samples these cycles at its own frame rate and phase, further affecting the recorded Glass-to-Glass latency distribution.

\begin{table}[!t]
\centering
\caption{Glass-to-Glass latency of RobotVision-PC and TeleMultiStream}
\label{tab:e2e-summary}
\small
\setlength{\tabcolsep}{2.5pt}
\renewcommand{\arraystretch}{1.08}
\begin{tabularx}{\columnwidth}{@{}Yrrr@{}}
\toprule
\multicolumn{4}{@{}l}{\textit{(a) Dominant interval}} \\
\textbf{Test condition} & \textbf{Range} & \textbf{Cov.} & \textbf{Median} \\
\midrule
\multicolumn{4}{@{}l}{\textbf{RobotVision-PC}~\cite{robotvision} \textit{(reference)}} \\
\mbox{Common ZED-M / SW / 4~Ultra} & 66--84 & 89.4\% & \textbf{82} \\
\midrule
\multicolumn{4}{@{}l}{\textbf{TeleMultiStream} \textit{(this work)}} \\
\mbox{Common ZED-M / SW / 4~Ultra} & 58--64 & 92.1\% & \textbf{60} \\
\mbox{Common ZED-M / HW / 4~Ultra} & 70--90 & 82.4\% & \textbf{80} \\
\mbox{Common ZED-M / HW-HP / 4~Ultra} & 60--70 & 86.2\% & \textbf{66} \\
\addlinespace[2pt]
\mbox{Config.~1 / SW / 4~Ultra} & 68--80 & 80.6\% & \textbf{72} \\
\mbox{Config.~1 / HW / 4~Ultra} & 86--102 & 81.2\% & \textbf{98} \\
\mbox{Config.~1 / HW-HP / 4~Ultra} & 62--74 & 82.7\% & \textbf{66} \\
\addlinespace[2pt]
\mbox{Config.~2 / SW / 4~Ultra} & 80--106 & 84.4\% & \textbf{94} \\
\mbox{Config.~2 / HW / 4~Ultra} & 100--130 & 84.1\% & \textbf{116} \\
\mbox{Config.~2 / HW-HP / 4~Ultra} & 68--100 & 80.0\% & \textbf{88} \\
\mbox{Config.~2 / SW / Swan} & 82--112 & 80.6\% & \textbf{102} \\
\addlinespace[2pt]
\mbox{Hikvision / SW / 4~Ultra} & 270--315 & 81.7\% & \textbf{280} \\
\mbox{Hikvision / SW / Swan} & 250--315 & 85.1\% & \textbf{280} \\
\bottomrule
\end{tabularx}
\par\vspace{5pt}
\begin{tabularx}{\columnwidth}{@{}Yrr@{}}
\toprule
\multicolumn{3}{@{}l}{\textit{(b) All samples}} \\
\textbf{Test condition} & \textbf{Range} & \textbf{Mean} \\
\midrule
\multicolumn{3}{@{}l}{\textbf{RobotVision-PC}~\cite{robotvision} \textit{(reference)}} \\
\mbox{Common ZED-M / SW / 4~Ultra} & 62--94 & 77.28 \\
\midrule
\multicolumn{3}{@{}l}{\textbf{TeleMultiStream} \textit{(this work)}} \\
\mbox{Common ZED-M / SW / 4~Ultra} & 58--68 & 61.21 \\
\mbox{Common ZED-M / HW / 4~Ultra} & 70--102 & 84.47 \\
\mbox{Common ZED-M / HW-HP / 4~Ultra} & 48--70 & 63.69 \\
\addlinespace[2pt]
\mbox{Config.~1 / SW / 4~Ultra} & 64--82 & 72.39 \\
\mbox{Config.~1 / HW / 4~Ultra} & 66--110 & 93.54 \\
\mbox{Config.~1 / HW-HP / 4~Ultra} & 60--84 & 68.40 \\
\addlinespace[2pt]
\mbox{Config.~2 / SW / 4~Ultra} & 80--120 & 96.58 \\
\mbox{Config.~2 / HW / 4~Ultra} & 82--142 & 114.78 \\
\mbox{Config.~2 / HW-HP / 4~Ultra} & 64--118 & 89.34 \\
\mbox{Config.~2 / SW / Swan} & 60--126 & 102.28 \\
\addlinespace[2pt]
\mbox{Hikvision / SW / 4~Ultra} & 245--350 & 285.65 \\
\mbox{Hikvision / SW / Swan} & 240--355 & 289.26 \\
\bottomrule
\end{tabularx}
\par\vspace{3pt}
\begin{minipage}{\columnwidth}
\footnotesize Latency in ms. Cov.: dominant-interval coverage.
\end{minipage}
\end{table}

\subsection{Common-Source Sender Comparison}

The common-source tests used the same ZED-M input, host, and network. TeleMultiStream SW and HW-HP achieved dominant-interval medians of 60 and 66~ms, respectively, versus 82~ms for RobotVision-PC, reducing latency by 26.8\% and 19.5\%. Baseline HW yielded 80~ms. SW and HW-HP therefore provided comparable visual latency in this comparison of complete implementations.

\begin{figure}[!htbp]
\centering
\resizebox{0.98\columnwidth}{!}{\input{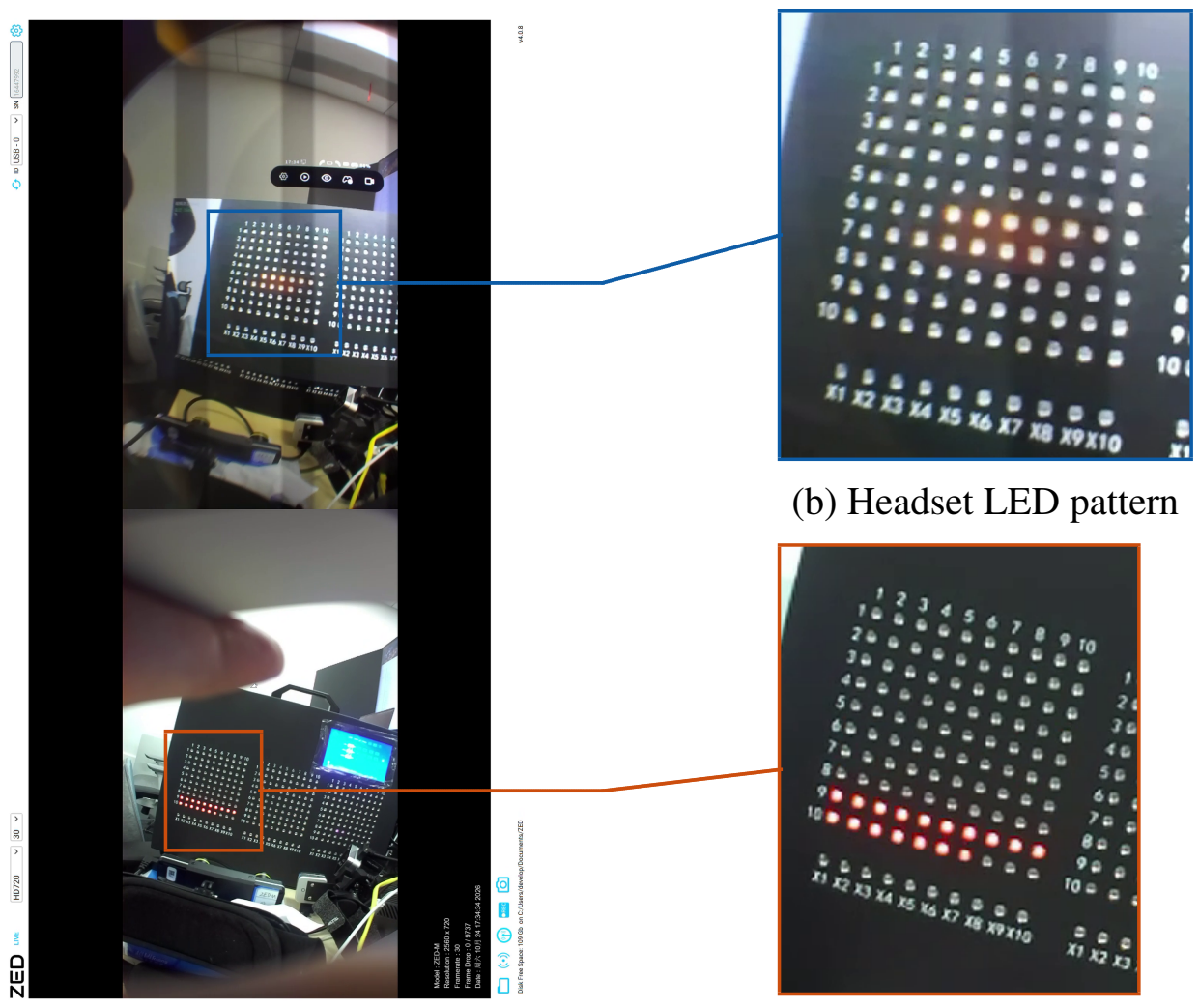}}
\caption{Glass-to-Glass measurement for Streaming Service with the common ZED-M source and software encoding. (a) The complete original recorded frame contains both views. The outlined regions are enlarged in (b) the headset LED pattern and (c) the directly observed LED pattern. At 2-ms LED increments, this capture corresponds to 60~ms.}
\label{fig:g2g-capture}
\end{figure}

Fig.~\ref{fig:g2g-capture} shows a common-source SW measurement: the complete recorded frame and linked enlargements preserve both LED views, whose offset at 2-ms increments corresponds to 60~ms.

Table~\ref{tab:zed-local} compares local stages across encoding modes. The SDK-to-Surface interval includes transport and buffering; camera acquisition and post-Surface display remain outside its boundaries (Section~\ref{sec:video-timing}).

\begin{table}[!t]
\centering
\caption{Common ZED-M source: mean local-stage durations (ms)}
\label{tab:zed-local}
\small
\setlength{\tabcolsep}{3pt}
\renewcommand{\arraystretch}{1.08}
\newcommand{\stagepoint}[1]{\begingroup\fontencoding{OMS}\fontfamily{cmsy}\fontsize{11}{11}\selectfont\textcircled{\normalfont\fontsize{6}{6}\selectfont #1}\endgroup}
\newcommand{\stagepair}[2]{\mbox{\stagepoint{#1}\,{\fontsize{9}{9}\selectfont$\rightarrow$}\,\stagepoint{#2}}}
\begin{tabularx}{\columnwidth}{@{}Yrrr@{}}
\toprule
\multirow{2}{*}{\textbf{Stage / boundary}} & \multicolumn{3}{c}{\textbf{TX encoder}} \\
\cmidrule(l){2-4}
 & \textbf{SW} & \textbf{HW} & \textbf{HW-HP} \\
\midrule
\multicolumn{4}{@{}l}{\textit{Sender (TX)}} \\
Raw-frame queueing \stagepair{3}{4} & 0.018 & 8.398 & 0.014 \\
Encoding time \stagepair{4}{5} & 3.782 & 19.888 & 4.039 \\
\quad VIC conversion & -- & 5.704 & 1.993 \\
\quad NVENC encoding & -- & 14.090 & 1.966 \\
\addlinespace[2pt]
\textbf{Sender processing time}\textsuperscript{a} & 6.367 & 30.541 & 6.303 \\
\midrule
\multicolumn{4}{@{}l}{\textit{Receiver (RX)}} \\
Decoding time \stagepair{7}{8} & 6.574 & 6.272 & 6.958 \\
Frame handoff \stagepair{8}{9} & 0.430 & 0.594 & 0.457 \\
Render queueing \stagepair{9}{10} & 0.759 & 0.869 & 0.918 \\
CPU drawing \stagepair{10}{11} & 0.215 & 0.227 & 0.229 \\
Surface swap \stagepair{12}{13} & 0.526 & 0.555 & 0.576 \\
\bottomrule
\end{tabularx}
\par\vspace{3pt}
\begin{minipage}{\columnwidth}
\footnotesize \textsuperscript{a}Composite interval defined in Table~\ref{tab:sender-intervals}.
\end{minipage}
\end{table}

HW-HP reduced mean encoding time from 19.89 to 4.04~ms (79.7\%), approaching SW's 3.78~ms. VIC conversion fell from 5.70 to 1.99~ms and NVENC encoding from 14.09 to 1.97~ms; raw-frame queueing fell from 8.40 to 0.014~ms. Receiver decoding remained similar across modes, locating the improvement primarily in the sender. HW-HP sustained approximately 59.84 Surface submissions per second.

Beyond encoding and raw-frame queueing, the main remaining costs are post-capture image retrieval and copying, plus WebRTC send preparation and transport submission. Larger frames increase retrieval work, while higher bitrates increase the encoded payload handled by WebRTC. These costs also explain the gaps in Tables~\ref{tab:c1-local}--\ref{tab:config-scale}.

\subsection{Encoding Modes in Configuration 1}

Configuration~1 ran three concurrent streams. The dominant-interval Glass-to-Glass medians were 72~ms for SW, 98~ms for HW, and 66~ms for HW-HP. Across the three HW-HP tests, the pooled full-sample mean was 68.40~ms, compared with 72.39~ms for SW. SW and HW-HP therefore provided comparable typical visual latency at this resolution. The HW-HP dominant-interval median was 6~ms lower.

\begin{table}[!t]
\centering
\caption{Mean ZED-M local-stage durations in Configuration 1 (ms)}
\label{tab:c1-local}
\small
\setlength{\tabcolsep}{3pt}
\renewcommand{\arraystretch}{1.08}
\newcommand{\stagepoint}[1]{\begingroup\fontencoding{OMS}\fontfamily{cmsy}\fontsize{11}{11}\selectfont\textcircled{\normalfont\fontsize{6}{6}\selectfont #1}\endgroup}
\newcommand{\stagepair}[2]{\mbox{\stagepoint{#1}\,{\fontsize{9}{9}\selectfont$\rightarrow$}\,\stagepoint{#2}}}
\begin{tabularx}{\columnwidth}{@{}Yrrr@{}}
\toprule
\multirow{2}{*}{\textbf{Stage / boundary}} & \multicolumn{3}{c}{\textbf{TX encoder}} \\
\cmidrule(l){2-4}
 & \textbf{SW} & \textbf{HW} & \textbf{HW-HP} \\
\midrule
\multicolumn{4}{@{}l}{\textit{Sender (TX)}} \\
Encoding time \stagepair{4}{5} & 4.711 & 20.340 & 4.362 \\
\quad VIC conversion & -- & 5.730 & 2.165 \\
\quad NVENC encoding & -- & 14.503 & 2.091 \\
\quad Encoder output wait (within \stagepair{4}{5}) & 0.001 & 20.316 & 4.337 \\
\addlinespace[2pt]
\textbf{Sender processing time}\textsuperscript{a} & 8.807 & 23.853 & 8.055 \\
\midrule
\multicolumn{4}{@{}l}{\textit{Receiver (RX)}} \\
Decoding time \stagepair{7}{8} & 5.787 & 6.350 & 5.942 \\
Frame handoff \stagepair{8}{9} & 0.350 & 0.383 & 0.364 \\
Render queueing \stagepair{9}{10} & 0.608 & 0.623 & 0.603 \\
CPU drawing \stagepair{10}{11} & 0.189 & 0.198 & 0.190 \\
Surface swap \stagepair{12}{13} & 0.430 & 0.412 & 0.426 \\
\bottomrule
\end{tabularx}
\par\vspace{3pt}
\begin{minipage}{\columnwidth}
\footnotesize Output wait overlaps conversion and encoding. \textsuperscript{a}Composite interval defined in Table~\ref{tab:sender-intervals}.
\end{minipage}
\end{table}

For ZED-M, mean encoding time was 4.71~ms for SW, 20.34~ms for HW, and 4.36~ms for HW-HP (Table~\ref{tab:c1-local}). Relative to HW, HW-HP reduced VIC conversion from 5.73 to 2.16~ms. It also reduced NVENC encoding from 14.50 to 2.09~ms. Together, these changes reduced the encoding total by 78.6\%. The two principal HW-HP stages account for approximately 97.6\% of that total. Under HW-HP, the two RealSense streams averaged 1.48 and 1.51~ms for encoding. All three streams sustained approximately 30 Surface submissions per second. The frequency optimization largely removed the baseline hardware-processing penalty while preserving low CPU utilization.

\subsection{Encoding Under Increased Stream Load}

Configuration~2 increased the ZED-M pixel count by a factor of 2.25 relative to Configuration~1. Each RealSense frame contained three times as many pixels. Target bitrates rose from 8 to 30~Mbps for ZED-M. They also rose from 4 to 8~Mbps for each RealSense stream. Thus, this comparison changes both resolution and bitrate.

\begin{table}[!t]
\centering
\caption{Mean ZED-M local-stage durations in Configuration 2 (ms)}
\label{tab:c2-local}
\small
\setlength{\tabcolsep}{3pt}
\renewcommand{\arraystretch}{1.08}
\newcommand{\stagepoint}[1]{\begingroup\fontencoding{OMS}\fontfamily{cmsy}\fontsize{11}{11}\selectfont\textcircled{\normalfont\fontsize{6}{6}\selectfont #1}\endgroup}
\newcommand{\stagepair}[2]{\mbox{\stagepoint{#1}\,{\fontsize{9}{9}\selectfont$\rightarrow$}\,\stagepoint{#2}}}
\begin{tabularx}{\columnwidth}{@{}Yrrr@{}}
\toprule
\multirow{2}{*}{\textbf{Stage / boundary}} & \multicolumn{3}{c}{\textbf{TX encoder}} \\
\cmidrule(l){2-4}
 & \textbf{SW} & \textbf{HW} & \textbf{HW-HP} \\
\midrule
\multicolumn{4}{@{}l}{\textit{Sender (TX)}} \\
Encoding time \stagepair{4}{5} & 11.854 & 31.040 & 8.802 \\
\quad VIC conversion & -- & 12.329 & 4.147 \\
\quad NVENC encoding & -- & 18.541 & 4.496 \\
\quad Encoder output wait & 0.001 & 31.004 & 8.766 \\
\addlinespace[2pt]
\textbf{Sender processing time}\textsuperscript{a} & 21.271 & 40.890 & 18.121 \\
\midrule
\multicolumn{4}{@{}l}{\textit{Receiver (RX)}} \\
Decoding time \stagepair{7}{8} & 9.296 & 9.615 & 8.946 \\
Frame handoff \stagepair{8}{9} & 0.394 & 0.417 & 0.371 \\
Render queueing \stagepair{9}{10} & 0.657 & 0.649 & 0.624 \\
CPU drawing \stagepair{10}{11} & 0.203 & 0.203 & 0.195 \\
Surface swap \stagepair{12}{13} & 0.441 & 0.455 & 0.425 \\
\bottomrule
\end{tabularx}
\par\vspace{3pt}
\begin{minipage}{\columnwidth}
\footnotesize Output wait overlaps conversion and encoding. \textsuperscript{a}Composite interval defined in Table~\ref{tab:sender-intervals}.
\end{minipage}
\end{table}

Table~\ref{tab:c2-local} compares the three encoding modes in Configuration~2. ZED-M encoding time averaged 11.85~ms for SW, 31.04~ms for HW, and 8.80~ms for HW-HP. Relative to HW, HW-HP reduced the encoding total by 22.24~ms (71.6\%). VIC conversion decreased from 12.33 to 4.15~ms, and NVENC encoding decreased from 18.54 to 4.50~ms. These two stages account for approximately 98.2\% of the HW-HP encoding total. Sender processing time decreased from 40.89 to 18.12~ms. Receiver decoding decreased from 9.61 to 8.95~ms. The measured improvement was therefore concentrated in the sender.

The dominant-interval Glass-to-Glass medians were 94~ms for SW, 116~ms for HW, and 88~ms for HW-HP. HW-HP reduced this median by 6.4\% relative to SW and by 24.1\% relative to HW. The corresponding full-sample means were 96.58, 114.78, and 89.34~ms. Full-sample medians were 98~ms for SW and 88~ms for HW-HP. Both hardware modes sustained approximately 29.87 Surface submissions per second for ZED-M and approximately 30 for each RealSense stream. CPU utilization was approximately 265\% for SW and 67\% for HW-HP.

\begin{table}[!t]
\centering
\caption{Mean ZED-M local-stage durations across configurations (ms)}
\label{tab:config-scale}
\small
\setlength{\tabcolsep}{2pt}
\renewcommand{\arraystretch}{1.08}
\newcommand{\stagepoint}[1]{\begingroup\fontencoding{OMS}\fontfamily{cmsy}\fontsize{11}{11}\selectfont\textcircled{\normalfont\fontsize{6}{6}\selectfont #1}\endgroup}
\newcommand{\stagepair}[2]{\mbox{\stagepoint{#1}\,{\fontsize{9}{9}\selectfont$\rightarrow$}\,\stagepoint{#2}}}
\begin{tabularx}{\columnwidth}{@{}Yrrrr@{}}
\toprule
\multirow{2}{*}{\textbf{Stage / boundary}} & \multicolumn{2}{c}{\textbf{Config. 1}} & \multicolumn{2}{c}{\textbf{Config. 2}} \\
\cmidrule(lr){2-3}\cmidrule(l){4-5}
 & \textbf{SW} & \textbf{HW-HP} & \textbf{SW} & \textbf{HW-HP} \\
\midrule
\multicolumn{5}{@{}l}{\textit{Sender (TX)}} \\
Source read \stagepair{1}{2} & 33.317 & 33.316 & 33.517 & 33.459 \\
Encoding time \stagepair{4}{5} & 4.711 & 4.362 & 11.854 & 8.802 \\
\quad VIC conversion & -- & 2.165 & -- & 4.147 \\
\quad NVENC encoding & -- & 2.091 & -- & 4.496 \\
\quad Encoder output wait & 0.001 & 4.337 & 0.001 & 8.766 \\
\addlinespace[2pt]
\textbf{Sender processing time}\textsuperscript{a} & 8.807 & 8.055 & 21.271 & 18.121 \\
\midrule
\multicolumn{5}{@{}l}{\textit{Receiver (RX)}} \\
Decoding time \stagepair{7}{8} & 5.787 & 5.942 & 9.296 & 8.946 \\
Frame handoff \stagepair{8}{9} & 0.350 & 0.364 & 0.394 & 0.371 \\
Render queueing \stagepair{9}{10} & 0.608 & 0.603 & 0.657 & 0.624 \\
CPU drawing \stagepair{10}{11} & 0.189 & 0.190 & 0.203 & 0.195 \\
Surface swap \stagepair{12}{13} & 0.430 & 0.426 & 0.441 & 0.425 \\
\bottomrule
\end{tabularx}
\par\vspace{3pt}
\begin{minipage}{\columnwidth}
\footnotesize Output wait overlaps conversion and encoding. \textsuperscript{a}Composite interval defined in Table~\ref{tab:sender-intervals}.
\end{minipage}
\end{table}

Table~\ref{tab:config-scale} compares SW and HW-HP across the two stream loads. On PICO 4 Ultra, the SW dominant-interval median rose from 72 to 94~ms. The HW-HP median rose from 66 to 88~ms. At the higher load, HW-HP reduced encoding time relative to SW by 25.7\%. Sender processing time decreased from 21.27 to 18.12~ms, while receiver decoding decreased from 9.30 to 8.95~ms.

\subsection{Headset Comparison in Configuration 2}\label{sec:config2-headsets}

Both headsets were tested with the same Configuration~2 source settings and software encoding. The dominant-interval Glass-to-Glass median was 94~ms on PICO 4 Ultra and 102~ms on PICO Swan. The full-sample medians showed the same ordering, at 98 and 105~ms. Table~\ref{tab:e2e-summary} gives the corresponding intervals and coverage.

\begin{table}[!t]
\centering
\caption{Mean receiver-stage durations for the ZED-M stream in Configuration 2 (ms)}
\label{tab:config2-devices}
\small
\setlength{\tabcolsep}{3pt}
\renewcommand{\arraystretch}{1.08}
\newcommand{\stagepoint}[1]{\begingroup\fontencoding{OMS}\fontfamily{cmsy}\fontsize{11}{11}\selectfont\textcircled{\normalfont\fontsize{6}{6}\selectfont #1}\endgroup}
\newcommand{\stagepair}[2]{\mbox{\stagepoint{#1}\,{\fontsize{9}{9}\selectfont$\rightarrow$}\,\stagepoint{#2}}}
\begin{tabularx}{\columnwidth}{@{}Yrr@{}}
\toprule
\textbf{Stage / boundary} & \textbf{4 Ultra} & \textbf{Swan} \\
\midrule
Decoding time \stagepair{7}{8} & 9.296 & 9.890 \\
Frame handoff \stagepair{8}{9} & 0.394 & 0.407 \\
Render queueing \stagepair{9}{10} & 0.657 & 0.843 \\
CPU drawing \stagepair{10}{11} & 0.203 & 0.257 \\
Surface swap \stagepair{12}{13} & 0.441 & 0.634 \\
\bottomrule
\end{tabularx}

\end{table}

Mean decoding time was 9.30~ms on PICO 4 Ultra and 9.89~ms on PICO Swan (Table~\ref{tab:config2-devices}). Swan's frame handoff, render queueing, CPU drawing, and Surface swap took an additional 0.013, 0.187, 0.054, and 0.193~ms, respectively. Mean sender encoding time differed by less than 0.1~ms between the two headsets. The measured local differences were small compared with the 8-ms difference in Glass-to-Glass medians. PICO Swan connects separate rendering and computation units through a wired interconnect. Transfer and synchronization between those units could increase the time from Surface-swap return to visible output relative to PICO 4 Ultra.

\subsection{High-Resolution Hikvision Input}

The Hikvision camera could only be connected to an x86 desktop in this setup. Its evaluation therefore covers x86 software encoding. Jetson hardware encoding was not tested. For the $8192\times3072$ side-by-side stereo input, both headsets had a dominant-interval Glass-to-Glass median of 280~ms. Full-sample means were 285.65~ms on PICO 4 Ultra and 289.26~ms on PICO Swan. The 3.61-ms difference was smaller than one 5-ms LED increment.

\begin{table}[!t]
\centering
\caption{Mean local-stage durations with Hikvision input (ms)}
\label{tab:hk-local}
\small
\setlength{\tabcolsep}{3pt}
\renewcommand{\arraystretch}{1.08}
\newcommand{\stagepoint}[1]{\begingroup\fontencoding{OMS}\fontfamily{cmsy}\fontsize{11}{11}\selectfont\textcircled{\normalfont\fontsize{6}{6}\selectfont #1}\endgroup}
\newcommand{\stagepair}[2]{\mbox{\stagepoint{#1}\,{\fontsize{9}{9}\selectfont$\rightarrow$}\,\stagepoint{#2}}}
\begin{tabularx}{\columnwidth}{@{}Yrr@{}}
\toprule
\textbf{Stage / boundary} & \textbf{4 Ultra} & \textbf{Swan} \\
\midrule
\multicolumn{3}{@{}l}{\textit{Sender (TX)}} \\
Encoding time \stagepair{4}{5} & 24.851 & 24.471 \\
\midrule
\multicolumn{3}{@{}l}{\textit{Receiver (RX)}} \\
Decoding time \stagepair{7}{8} & 50.472 & 37.075 \\
Frame handoff \stagepair{8}{9} & 26.203 & 27.593 \\
Render queueing \stagepair{9}{10} & 0.881 & 1.372 \\
CPU drawing \stagepair{10}{11} & 0.250 & 0.371 \\
Surface swap \stagepair{12}{13} & 0.550 & 0.913 \\
\bottomrule
\end{tabularx}

\end{table}

For successfully processed frames, receiver decoding time averaged 37.08~ms on Swan and 50.47~ms on 4 Ultra (Table~\ref{tab:hk-local}). The reduction was approximately 13.40~ms (26.5\%). The lower mean indicates faster hardware decoding on Swan under this workload.

Each other measured stage differed by less than 1.4~ms. Swan's shorter decoding time and the equal 280-ms Glass-to-Glass medians are consistent with additional time between Surface submission and visible output on Swan, as discussed in Section~\ref{sec:config2-headsets}.

\subsection{UDP Pose Downlink Under Concurrent Video Load}
\label{sec:udp-pose-results}

While receiving three streams in Configurations~1 and~2, the PICO application sent head and controller poses to Jetson Orin over UDP at 100~Hz. The sending interval was 10~ms. Each configuration was tested with SW, HW, and HW-HP, with five trials per condition.

Corrected one-way pose latency spans the PICO message timestamp to application reception on Orin. The PICO millisecond tick was mapped to the Orin monotonic clock by interpolation between calibration anchors, each averaged from 50 exchanges. Robot execution follows the reception endpoint. Negative corrected values are retained, and packets without calibration coverage are excluded from latency statistics. Packet loss is estimated from message-timestamp intervals and the configured sending frequency. Table~\ref{tab:udp-pose} reports the mean latency and mean estimated packet loss for each condition, with equal weight given to each trial. Clock-mapping error, timestamp quantization, and sending-rate variation remain measurement limitations.

\begin{table}[!t]
\centering
\caption{UDP pose downlink during concurrent three-stream video reception}
\label{tab:udp-pose}
\small
\setlength{\tabcolsep}{3pt}
\renewcommand{\arraystretch}{1.08}
\begin{tabularx}{\columnwidth}{@{}Yrrr@{}}
\toprule
\textbf{Condition} & \textbf{Packets} & \textbf{Latency (ms)} & \textbf{Loss (\%)} \\
\midrule
Config. 1 SW & 29,998 & 1.780 & 0.0067 \\
Config. 1 HW & 29,998 & 2.137 & 0.0067 \\
Config. 1 HW-HP & 30,000 & 1.576 & 0.0033 \\
Config. 2 SW & 29,970 & 2.532 & 0.1333 \\
Config. 2 HW & 29,986 & 3.074 & 0.0667 \\
Config. 2 HW-HP & 29,998 & 1.800 & 0.0033 \\
\bottomrule
\end{tabularx}
\par\vspace{3pt}
\begin{minipage}{\columnwidth}
\footnotesize Latency and estimated loss are means across the five trials per condition. Packets are total received datagrams.
\end{minipage}
\end{table}

HW-HP had the lowest mean corrected one-way pose latency in both configurations: 1.576~ms in Configuration~1 and 1.800~ms in Configuration~2, with mean estimated packet loss of 0.0033\% in each. These measurements support millisecond-scale pose delivery with low estimated loss during concurrent three-stream video reception.

\subsection{Latency, CPU Utilization, and Power Consumption}

Panel (a) of Table~\ref{tab:latency-cpu} summarizes Glass-to-Glass latency and sender CPU utilization for all three encoding conditions. HW and HW-HP had approximately equal CPU utilization at each load. Their values were 32\% for the common source, 38\% for Configuration~1, and 67\% for Configuration~2. SW used 96\%, 120\%, and 265\%, respectively. Thus, HW-HP retained the CPU advantage of hardware encoding. At the same time, it substantially reduced latency relative to baseline HW.

\begin{table}[!htbp]
\centering
\caption{Joint assessment of Glass-to-Glass latency, sender CPU utilization, and estimated module power}
\label{tab:latency-cpu}
\small
\setlength{\tabcolsep}{2pt}
\renewcommand{\arraystretch}{1.08}
\begin{tabularx}{\columnwidth}{@{}L{.25\columnwidth}*{6}{>{\raggedleft\arraybackslash}X}@{}}
\toprule
\multicolumn{7}{@{}l}{\textit{(a) Latency and sender CPU utilization}} \\
\multirow{2}{*}{\textbf{Setting}} & \multicolumn{3}{c}{\textbf{Glass-to-Glass (ms)}} & \multicolumn{3}{c}{\textbf{Sender CPU (\%)}} \\
\cmidrule(lr){2-4}\cmidrule(l){5-7}
 & \textbf{SW} & \textbf{HW} & \textbf{HP} & \textbf{SW} & \textbf{HW} & \textbf{HP} \\
\midrule
Common ZED-M & 60 & 80 & 66 & 96 & 32 & 32 \\
Config. 1 & 72 & 98 & 66 & 120 & 38 & 38 \\
Config. 2 & 94 & 116 & 88 & 265 & 67 & 67 \\
\bottomrule
\end{tabularx}
\par\vspace{5pt}
\begin{tabularx}{\columnwidth}{@{}Yrrr@{}}
\toprule
\multicolumn{4}{@{}l}{\textit{(b) Mean estimated module power (W)}} \\
\textbf{Setting} & \textbf{SW} & \textbf{HW} & \textbf{HW-HP} \\
\midrule
Common ZED-M & 14.703 & 13.942 & 15.420 \\
Config. 1 & 14.130 & 13.349 & 14.431 \\
Config. 2 & 17.524 & 15.498 & 16.581 \\
\bottomrule
\end{tabularx}

\par\vspace{3pt}
\begin{minipage}{\columnwidth}
\footnotesize HP denotes HW-HP. Glass-to-Glass values are dominant-interval medians. CPU denotes sender-process utilization. Estimated module power sums the GPU/SoC, CPU/CV, and system 5V rail means.
\end{minipage}
\end{table}

In all three settings, the measured HW-HP frequencies were 998~MHz for NVENC and 729~MHz for VIC. Mean power was calculated by averaging the instantaneous readings reported by \texttt{tegrastats}. Module power was estimated as the sum of the mean powers on the \texttt{VDD\_GPU\_SOC}, \texttt{VDD\_CPU\_CV}, and \texttt{VIN\_SYS\_5V0} rails, following NVIDIA's benchmark method~\cite{modulepower}. This estimate covers the Jetson module.

Panel (b) reports the estimated module power. Relative to HW, HW-HP increased it by 1.477~W for the common source and approximately 1.082~W for each multi-stream configuration. The dominant-interval Glass-to-Glass medians decreased by 14, 32, and 28~ms for the common source and Configurations~1 and~2, respectively.

Relative to SW, HW-HP increased estimated module power by 0.717~W (4.9\%) for the common source and 0.301~W (2.1\%) for Configuration~1. In Configuration~2, it reduced estimated module power from 17.524 to 16.581~W (5.4\%). The CPU/CV-rail power decreased from 4.786 to 2.393~W, offsetting the higher GPU/SoC-rail power.

At the lower tested resolutions, SW and HW-HP provided comparable typical visual latency. SW had slightly lower estimated module power but substantially higher CPU utilization. It remains a viable option when CPU resources are sufficient. In the higher-resolution multi-stream Configuration~2, HW-HP offered lower encoding time, typical visual latency, CPU utilization, and estimated module power.

\FloatBarrier

\section{Conclusions}\label{sec:conclusion}

Training embodied-intelligence models relies on large-scale, high-quality demonstration data. Robot teleoperation provides an important means of collecting these data. Low-latency feedback and control also support timely human intervention and corrective demonstrations in DAgger-based interactive learning. TeleMultiStream addresses these requirements by combining multi-stream WebRTC video feedback with UDP pose downlink. Its Streaming Service sender delivers head- and wrist-camera views to the native PICO receiver, while the downlink carries operator motion to the robot.

With the common ZED-M source on a wired LAN, the dominant-interval Glass-to-Glass median was 60~ms for software encoding and 66~ms for high-performance hardware encoding, compared with 82~ms for RobotVision-PC. Raising VIC and NVENC frequencies through device nodes improved the hardware path without changing sender or receiver source code. Software and high-performance hardware encoding produced comparable latency in the common-source and Configuration~1 tests. In Configuration~2, high-performance hardware encoding reduced encoding time by 25.7\% and the dominant-interval Glass-to-Glass median from 94 to 88~ms, while using approximately 67\% CPU versus 265\% for software encoding.

At the lower tested resolutions, software encoding had slightly lower estimated module power but substantially higher CPU utilization. It remains viable when CPU resources are sufficient. In Configuration~2, high-performance hardware encoding reduced estimated module power by 5.4\% relative to software encoding. Its lower latency and CPU utilization therefore favor this mode for the tested higher-resolution multi-stream workload. During three-stream reception with HW-HP at 100~Hz, mean corrected one-way pose latency was 1.576~ms in Configuration~1 and 1.800~ms in Configuration~2.

\begingroup\clubpenalty=10000
Common source and encoder interfaces allow TeleMultiStream to combine camera and ROS image inputs and select an encoding backend suited to the host. The sender can also publish the same encoded streams to ROS2 for recording. These capabilities provide reusable infrastructure for collecting multi-view robot demonstrations across camera layouts and CPU budgets.\par\endgroup

Future work will evaluate whether alternative camera interfaces can make captured frames available to the sender earlier.

\FloatBarrier
\IEEEtriggeratref{24}

\end{document}